\documentclass{article}

 \usepackage[preprint]{neurips_2025}

\usepackage[utf8]{inputenc} 
\usepackage[T1]{fontenc}    
\usepackage{hyperref}       
\usepackage{url}            
\usepackage{booktabs}       
\usepackage{amsfonts}       
\usepackage{nicefrac}       
\usepackage{microtype}      
\usepackage[table]{xcolor}  
\usepackage{times}
\usepackage{booktabs}
\usepackage{amsmath}
\usepackage{latexsym}
\usepackage{graphicx}
\usepackage{subcaption}
\usepackage{wrapfig}
\usepackage{multirow}
\usepackage{tabularx}

\usepackage{array}
\usepackage{enumitem}
\usepackage[most]{tcolorbox}
\usepackage{listings}

\definecolor{tracebg}{HTML}{F5F7FA}
\definecolor{traceframe}{HTML}{6B7C99}
\definecolor{promptbg}{HTML}{F8F4EC}
\definecolor{promptframe}{HTML}{B58A3F}
\newtcolorbox{tracebox}[1][]{
  enhanced, breakable,
  colback=tracebg, colframe=traceframe,
  boxrule=0.5pt, arc=2pt,
  left=6pt, right=6pt, top=4pt, bottom=4pt,
  fonttitle=\bfseries\small\sffamily,
  fontupper=\footnotesize\ttfamily,
  title={#1},
}
\newtcolorbox{promptbox}[1][]{
  enhanced, breakable,
  colback=promptbg, colframe=promptframe,
  boxrule=0.5pt, arc=2pt,
  left=6pt, right=6pt, top=4pt, bottom=4pt,
  fonttitle=\bfseries\small\sffamily,
  fontupper=\footnotesize\ttfamily,
  title={#1},
}
\lstdefinelanguage{json}{
  basicstyle=\ttfamily\small,
  numbers=none,
  stepnumber=1,
  numbersep=8pt,
  showstringspaces=false,
  breaklines=true,
  frame=single,
  backgroundcolor=\color{black!2},
  stringstyle=\color{green!50!black},
  keywordstyle=\color{blue!70},
  commentstyle=\color{black!60},
  morekeywords={true,false,null},
}
 \usepackage{amssymb}
 \definecolor{c1}{RGB}{239,118,122}
\definecolor{c2}{RGB}{69,105,144}
\definecolor{c3}{RGB}{72,192,170}
\definecolor{c4}{RGB}{179,149,189}
\usepackage[T1]{fontenc}
\title{AgentFugue: Agent Scaling for Long-Horizon Tasks through Collective Reasoning}

\author{Yuyang Hu$^{1,2}$\thanks{Equal contribution.}, Hongjin Qian$^2$\footnotemark[1], Shuting Wang$^1$, Jiongnan Liu$^1$, Tong Zhao$^1$, Xiaoxi Li$^1$ \\
\textbf{Zheng Liu}$^2$\thanks{Corresponding author.}, \textbf{Zhicheng Dou}$^1$\footnotemark[2] \\
        $^1$ GSAI, Renmin University of China \\ 
        $^2$ Beijing Academy of Artificial Intelligence \\ 
}

\begin{document}

\maketitle

\begin{abstract}
Recent progress on long-horizon agentic tasks has been driven largely by scaling up individual agents through stronger models, better tools, and more effective scaffolding. In contrast, much less is understood about scaling out: whether multiple peer agents, all targeting the same task, can become an additional source of capability without relying on explicit role specialization or workflow orchestration. We study this question and propose \textbf{AgentFugue}, a collective reasoning framework built around a shared reasoning hub. As peer agents explore the same task in parallel, the hub records concise notes on what each agent has established, attempted, or ruled out, and enables each agent to selectively access what other agents have discovered in a form useful for its current search. This design turns otherwise isolated trajectories into a connected ecology of reusable intermediate reasoning without requiring centralized planning. We instantiate the hub as a plug-in communication layer, trained with supervised fine-tuning and end-to-end reinforcement learning. Across the challenging long-horizon settings we study, AgentFugue improves over strong baselines. Our results suggest that collective reasoning can turn scaling out peer agent systems into a distinct source of capability gains, rather than merely a way of spending more compute. Our code is available at \url{https://github.com/qhjqhj00/cabeza}

\end{abstract}

\section{Introduction}
\label{sec:intro}

Recent progress has shown that LLM-based agents can perform remarkably well on complex long-horizon tasks~\citep{nakano2021webgpt,qiao2025webresearcher,chen2025iterresearch,ye2026agentfold,wu2025webdancer,zheng2025deepresearcher,jin2025searchr1}. A key driver of this progress is sustained scaling up along several dimensions, including stronger foundation models~\citep{openai2023gpt4,grattafiori2024llama3,guo2025deepseekr1,yang2025qwen3}, better tool use~\citep{schick2023toolformer,patil2023gorilla,qin2023toolllm,wang2024codeact,li2025searcho1,li2025webthinker,jin2025searchr1}, and more effective agent scaffolding~\citep{yao2023react,shinn2023reflexion,yao2023tot,wang2023voyager,jin2025flashrag}. This scaling-up paradigm has substantially expanded what a single agent can do. At the same time, however, it improves the strength of one trajectory rather than the breadth of exploration, leaving open the question of whether complex tasks may also benefit from scaling beyond a single agent.

Prior work has also shown that multi-agent systems can be effective for complex tasks~\citep{wu2023autogen,hong2023metagpt,li2023camel}, but the dominant emphasis has been on orchestration: assigning different roles to different agents~\citep{qian2023chatdev,huang2023agentcoder,chen2023agentverse}, decomposing tasks into separate subtasks~\citep{shen2023hugginggpt,wang2023plansolve,k2.5}, or designing explicit interaction workflows~\citep{zhuge2024gptswarm,liu2023dylan,qian2024macnet,zhang2024aflow}. A complementary line of work coordinates multiple agents through deliberation~\citep{du2023debate,liang2023mad,chen2023reconcile}. Such approaches improve capability through structured coordination, with different agents contributing in different ways. What remains less understood is whether gains can also arise in a simpler setting, where multiple agents act as peers on the same task rather than being separated by pre-defined responsibilities.

This peer setting creates a different opportunity for capability growth. When multiple agents explore the same task in parallel~\citep{wang2023selfconsistency,brown2024monkeys,snell2025scaling,li2024moreagents}, they may uncover different partial reasoning paths, intermediate evidence, or failed branches. We study whether such parallel exploration can itself become a source of additional capability, rather than merely additional compute. This is the sense in which we use the term \emph{scaling out}: increasing the number or diversity of peer agents working on the same task so that their trajectories can inform and redirect one another. Realizing this benefit, however, is non-trivial. Without communication, multiple agents largely reduce to isolated searches whose results must be merged after the fact~\citep{wang2023selfconsistency,brown2024monkeys,li2024moreagents,lee2026agenticaggregationparallelscaling}; with unrestricted communication, useful signals can be overwhelmed by raw trajectory noise~\citep{li2025parallelmuse}, and the diversity of exploration may quickly collapse. We therefore argue that effective scaling out requires a mechanism for \emph{collective reasoning}, through which peer agents can selectively exchange intermediate progress while continuing to explore the same task from different directions. In this sense, collective reasoning is best understood not as a shared conversation~\citep{du2023debate,liang2023mad,chen2023reconcile}, but as a fugue-like structure of parallel search: in the spirit of a Baroque fugue, multiple trajectories remain distinct while still picking up and developing one another's partial progress~\citep{mann1987study}.

To realize this form of collective reasoning, we propose \textbf{AgentFugue}, a framework built around a shared reasoning hub. The hub serves as an external communication layer rather than a centralized planner: when an agent completes a coherent episode of interaction, the hub records a compact note about what that agent established, attempted, or ruled out, and later allows other agents to selectively access the parts of that progress that are useful for their own search. Because the hub is attached outside the core policy, similar in spirit to externalized memory modules studied for single-agent settings~\citep{mem02025,fang2025lightmem,tan2026memsifter,xu2025amem,hu2026compassmem}, AgentFugue is adaptive to different reasoning agents while preserving the independence of their local trajectories.

This design lets us study two complementary forms of scaling out. In \emph{homogeneous teams}, multiple agents share the same backbone and configuration, so any gain must come from interaction among parallel trajectories rather than from built-in role differences. In \emph{heterogeneous teams}, agents differ in model or setup, making it possible for distinct reasoning biases to complement one another on the same task~\citep{wang2024moa}. Across both settings, the central empirical question is whether collective reasoning can improve not only team-level success, but also the quality and efficiency of the individual trajectories that make up the team.
In our implementation, the shared reasoning hub is optimized separately from the task agents themselves. We instantiate its write and read functions with a moderate-sized language model, then improve them through supervised fine-tuning followed by end-to-end reinforcement learning so that the hub learns not only to summarize intermediate progress, but also to return guidance that is useful inside the full agent loop.

We evaluate AgentFugue on challenging long-horizon benchmarks spanning information seeking, open-ended problem solving, and multi-step web reasoning. Across the settings we study, we observe gains in both homogeneous and heterogeneous teams, supporting the view that peer-agent communication can provide a robust source of capability beyond stronger individual agents alone.

Our contributions are threefold: (1) we identify peer-agent scaling as a distinct setting for long-horizon reasoning, in which multiple agents work on the same task and capability must arise from cross-trajectory reuse rather than role specialization. (2) we propose AgentFugue, a communication framework based on writing, retrieving, and reading shared reasoning episodes, which turns parallel trajectories into a selectively shared reasoning ecology without centralized planning. (3) we study this framework in both homogeneous and heterogeneous teams, with analyses designed to test when scaling out improves per-agent efficiency, when it yields larger team-level gains, and where the communication mechanism breaks down.

\section{Method}
\label{sec:method}

\begin{figure*}[t]
    \centering
    \includegraphics[width=0.98\textwidth]{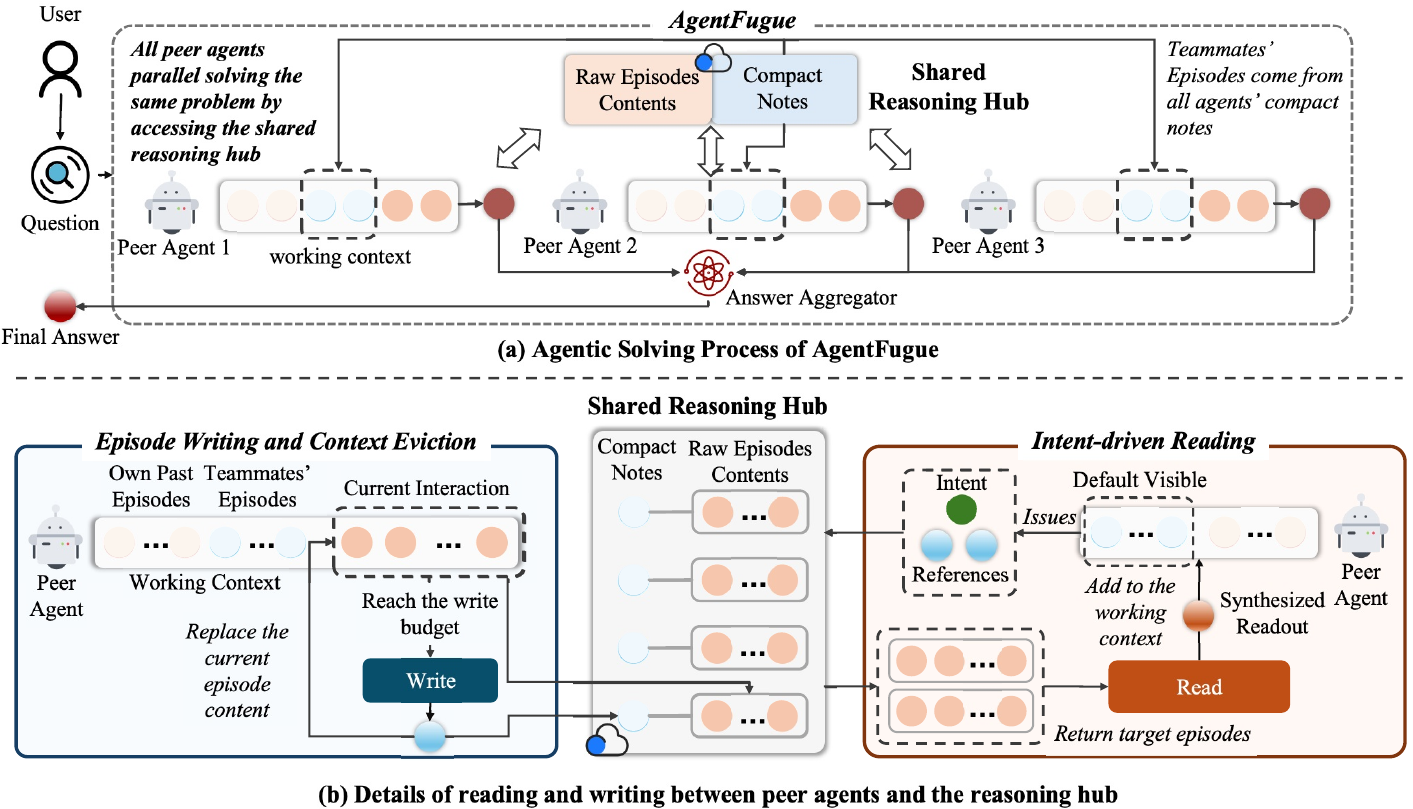}
    \caption{\textbf{Overview of AgentFugue.} The top panel illustrates the core idea: peer agents explore the same task in parallel while a shared reasoning hub mediates cross-trajectory communication. The bottom panel details the reasoning hub, including episode writing, context eviction, and intent-driven reading. Once an agent's current interaction reaches the write budget, it is summarized into an episode note and added to the hub; later, another agent can consult relevant teammate episodes and receive synthesized guidance for its ongoing trajectory.}
    \label{fig:framework}
\end{figure*}

\subsection{Problem Setting}
\label{sec:knowledge-space}

\paragraph{Target knowledge space.}
Consider a long-horizon task instance $x$ whose solution requires assembling a body of evidence and reasoning that we call the \emph{target knowledge space} $\mathcal{K}^*(x)$. For hard tasks, $\mathcal{K}^*(x)$ is large and structurally complex: it may span multiple evidence types, reasoning chains, and verification steps. In a single-agent run, the agent explores a trajectory $\tau$ and accumulates a discovered subspace $\mathcal{K}(\tau) \subseteq \mathcal{K}^*(x)$. Any single trajectory is unlikely to cover $\mathcal{K}^*(x)$ fully, since each run touches a different, partial, and often sub-optimal fragment, partly by skill and partly by the luck of which branches happen to be explored. Scaling out, by running multiple peer agents on the same task, creates the opportunity for their discovered fragments to complement one another, but only if the fragments can be shared.

\paragraph{Task, team, and trajectories.}
We formalize this setting as follows. A team of $N$ agents all target the same task instance $x$. Agent $i \in \{1,\ldots,N\}$ interacts with the environment through reasoning steps, tool calls, and observations, producing a local trajectory whose prefix up to step $t$ is
\begin{equation}
\tau_{i,t} = \big[(a_{i,1},o_{i,1}), \ldots, (a_{i,t},o_{i,t})\big],
\end{equation}
where $a_{i,t}$ is an action and $o_{i,t}$ the resulting observation. Each trajectory represents a different exploratory path through $\mathcal{K}^*(x)$, with its own discovered subspace $\mathcal{K}(\tau_i)$.

\paragraph{Shared reasoning hub.}
To connect these scattered fragments, the team is augmented with a \emph{shared reasoning hub} $\mathcal{H}$. As shown in Figure~\ref{fig:framework}, the hub sits alongside the peer agents as a team-level communication interface: it compresses completed portions of each agent's reasoning history into reusable notes and allows agents to consult one another's progress during search. Its role is not to replace local reasoning or to centrally orchestrate the team, but to make intermediate discoveries produced by one trajectory selectively available to others, thereby expanding each agent's effective knowledge space beyond what its own trajectory covers.

\paragraph{Episodes.}
To make partial progress shareable, we divide each local trajectory into completed \emph{episodes}. An episode $\epsilon_{i,e}$ is a contiguous chunk of interaction history determined by a fixed local context budget:
\begin{equation}
\epsilon_{i,e} = \big[(a_{i,t_s},o_{i,t_s}), \ldots, (a_{i,t_e},o_{i,t_e})\big].
\end{equation}
Once the active context reaches the budget, the accumulated segment is summarized and written to the hub. Episodes are therefore the units through which an agent's partial progress becomes visible to the rest of the team through $\mathcal{H}$. At any point during search, an agent may either continue along its own local trajectory or consult $\mathcal{H}$ to access relevant progress produced by other team members.

This formulation subsumes several useful limiting cases. When $N{=}1$, it reduces to a single reasoning agent coupled with an external memory-like module. When $N{>}1$ but no agent consults the hub, it reduces to multiple isolated trajectories that share compute but not information. Our main interest lies between these extremes: peer-agent teams in which multiple agents pursue the same task while selectively reusing one another's intermediate reasoning.

\paragraph{Two forms of scaling out.}
We study this setting in two forms. In \emph{homogeneous} teams, all agents share the same model and configuration, so any gains must arise from cross-trajectory interaction rather than built-in agent differences. In \emph{heterogeneous} teams, agents differ in model backbone or prompting configuration, introducing systematic diversity beyond stochastic variation: different models carry different reasoning biases, knowledge distributions, and failure modes, so the hub can additionally mediate complementary strengths across the team.

\paragraph{From isolated fragments to connected knowledge.}
Without communication, the team's collective knowledge $\bigcup_i \mathcal{K}(\tau_i)$ exists only in aggregate: no individual agent can access another's discoveries, so each remains limited to its own fragment. The role of $\mathcal{H}$ is to connect these scattered fragments by making useful portions of one trajectory selectively available to another, expanding each agent's effective knowledge space beyond $\mathcal{K}(\tau_{i,t})$ alone. This perspective clarifies both the promise and the limit of scaling out: adding agents increases the diversity of discovered fragments, but the marginal gain depends on whether new trajectories reach genuinely new regions of the task-relevant knowledge needed to solve $x$, denoted conceptually by $\mathcal{K}^*(x)$, and whether the hub can surface those regions when they are needed. The rest of this section describes the hub mechanism that operationalizes this view (\S\ref{sec:hub}) and how we optimize it (\S\ref{sec:hub-opt}).

\subsection{Shared Reasoning Hub}
\label{sec:hub}
AgentFugue operationalizes the shared reasoning hub through two operations: \emph{episode writing}, which compresses completed trajectory segments into reusable notes, and \emph{intent-driven reading}, which lets agents inspect and synthesize relevant teammate episodes on demand.

\paragraph{Episode writing and context eviction.}
As illustrated in the top-right panel of Figure~\ref{fig:framework}, agent $i$'s local context window accumulates reasoning steps, tool calls, and observations until it reaches a fixed write budget, at which point the current segment is closed as an episode $\epsilon_{i,e}$. The hub model then compresses the episode into an \emph{episode note}:
\begin{equation}
z_{i,e} = M_{\mathrm{write}}(\epsilon_{i,e}),
\end{equation}
which captures the team-relevant content of that episode: what was established, what evidence was collected, what was attempted, and which branches were ruled out. Once the note is written, the raw episode content in the agent's working context is \emph{evicted and replaced} by its episode note $z_{i,e}$. This serves a dual purpose: it compresses the agent's own history to free context capacity for continued exploration, and it produces a representation suitable for sharing with other agents. The full episode content $\epsilon_{i,e}$ is retained in the hub's storage for later deep reading.

At any point during search, agent $i$'s working context therefore takes the form:
\begin{equation}
\label{eq:context}
\mathcal{C}_{i,t} = \big[\,\underbrace{z_{i,1},\ldots,z_{i,e-1}}_{\text{own episode notes}},\;\; \underbrace{z_{j_1, e'},\ldots,z_{j_k, e''}}_{\text{teammates' notes}},\;\; \underbrace{\tau_{i,t}^{\mathrm{active}}}_{\text{current interaction}}\,\big],
\end{equation}
where the first group contains episode notes summarizing agent $i$'s own completed episodes, the second group contains episode notes from other agents that have been made visible through prior hub interactions, and $\tau_{i,t}^{\mathrm{active}}$ is the current unfinished interaction segment. This design keeps the working context bounded even as total reasoning effort grows, while exposing a structured view of the team's collective progress.

\paragraph{Intent-driven reading.}
As shown in the bottom-right panel of Figure~\ref{fig:framework}, agents do not passively receive all teammate episode notes. Instead, when agent $i$ judges, based on its current context $\mathcal{C}_{i,t}$, that consulting a teammate's work in greater depth would be useful, it issues a structured request to the hub with two components: an \emph{intent} $q_{i,t}$ describing what kind of information is needed, and a set of \emph{episode references} $\mathcal{E}_{i,t} \subseteq \{\epsilon_{j,e} \mid j \neq i\}$ indicating which teammates' episodes it wants to inspect in full. The agent selects these references based on the episode notes already visible in $\mathcal{C}_{i,t}$: for example, an episode note may indicate that another agent found evidence related to the current search direction, prompting a request for the original episode.

Given this request, the hub retrieves the full raw content of the referenced episodes from its storage and synthesizes them in light of the intent:
\begin{equation}
r_{i,t} = M_{\mathrm{read}}\!\big(q_{i,t},\;\; \{\epsilon_{j,e} : \epsilon_{j,e} \in \mathcal{E}_{i,t}\}\big).
\end{equation}
The resulting readout $r_{i,t}$ is a focused piece of evidence or guidance tailored to the requesting agent's current need, which is appended to $\mathcal{C}_{i,t}$. In this design, episode notes provide coarse awareness and help the agent identify which episodes are worth inspecting, while the hub performs the actual synthesis over the raw referenced content. This two-level design, with episode notes for broad awareness and intent-driven reading for selective depth, avoids both extremes of no communication and full broadcast. Agents maintain a lightweight overview of team progress through episode notes and can drill into specific episodes when deeper information is needed.

\paragraph{Distinction from nearby paradigms.}
The write/read mechanism differs from several adjacent settings in important ways. Unlike single-agent memory, notes are written to support cross-agent reuse, not just the originating trajectory. Unlike multi-agent debate or group chat, agents are not forced into synchronized turn-taking or a shared conversational context. Unlike best-of-$N$ sampling, trajectories influence one another \emph{before} completion through reusable intermediate progress. And unlike RAG-style retrieval over a static corpus, the read path is intent-driven and synthesizes \emph{raw episode content} on demand rather than returning pre-formed passages.

\subsection{Hub Optimization}
\label{sec:hub-opt}
The hub is initialized from a Qwen3.5-9B backbone, with separate $M_{\mathrm{write}}$/$M_{\mathrm{read}}$ instances from the same checkpoint, and optimized in two stages.

\paragraph{Supervised fine-tuning.}
A teacher model produces reference notes $z^*$ for each completed episode and reference readouts $r^*$ for each read request, yielding $\mathcal{D}_{\mathrm{write}}$ and $\mathcal{D}_{\mathrm{read}}$. Both heads are trained jointly with the standard LM loss:
\begin{equation}
\mathcal{L}_{\mathrm{SFT}} = -\mathbb{E}_{\mathcal{D}_{\mathrm{write}}}\!\big[\log \pi_\theta(z^* \!\mid\! \epsilon)\big] - \mathbb{E}_{\mathcal{D}_{\mathrm{read}}}\!\big[\log \pi_\theta(r^* \!\mid\! q, \mathcal{E})\big].
\end{equation}

\paragraph{Group relative policy optimization.}
We then align the hub with downstream task success via GRPO in the full multi-agent loop, keeping task agents frozen. For instance $x$, sample $G$ candidate hub outputs $\{y_g\}$, run the loop with each, observe rewards $\{R_g\}$, form group-relative advantages $\hat{A}_g = (R_g - \mathrm{mean})/(\mathrm{std} + \varepsilon)$, and optimize:
\begin{equation}
\mathcal{J}_{\mathrm{GRPO}} = \mathbb{E}\!\Big[\tfrac{1}{G}\!\sum_{g} \min\!\big(\rho_g\hat{A}_g,\,\mathrm{clip}(\rho_g, 1{-}\delta, 1{+}\delta)\hat{A}_g\big)\Big] - \beta\, D_{\mathrm{KL}}\!\big[\pi_\theta \,\|\, \pi_{\mathrm{ref}}\big],
\end{equation}
with $\rho_g = \pi_\theta(y_g)/\pi_{\mathrm{old}}(y_g)$, $\pi_{\mathrm{ref}}$ the SFT checkpoint, and $R_g$ combining task success with a brevity bonus that favors hub outputs leading to shorter effective search paths. Because task agents are frozen, GRPO pressure lands on the communication layer itself.

\section{Experiments}
\label{sec:exp}

\subsection{Datasets}

We evaluate on three benchmarks chosen to stress complementary aspects of long-horizon agentic reasoning: \textbf{BrowseComp}~\citep{browsecomp}, which requires deep multi-hop web search and cross-document evidence aggregation toward a short factual answer; \textbf{WideSearch}~\citep{widesearch}, which rewards \emph{breadth} of evidence collection rather than depth, asking agents to enumerate and consolidate many parallel pieces of information; and \textbf{HLE (Humanity's Last Exam)}~\citep{hle}, an expert-authored multi-domain reasoning benchmark whose questions stress deliberate multi-step reasoning rather than web navigation. For all three benchmarks we follow the official judging protocol. For evaluation efficiency and to keep the compute budget manageable, on BrowseComp and HLE we follow prior work~\citep{li2025webthinker,lee2026agenticaggregationparallelscaling,feng2026agentswing} and evaluate on a $200$-question random sample rather than the full test set; WideSearch is used in full. More details are deferred to Appendix~\ref{app:datasets}.

\subsection{Baselines}

We compare against three groups of systems (Appendix~\ref{app:baselines}); all multi-agent systems share the same per-agent tool stack and interaction budget so that any difference reflects coordination, not capability.

\paragraph{Single-agent ReAct.} Frontier models in a standard ReAct~\citep{yao2023react} loop with the same tool stack as AgentFugue, isolating how far ``scaling up'' a single agent goes: \textbf{Claude-Opus-4.5}, \textbf{Kimi-K2.5}~\citep{k2.5}, \textbf{Qwen3.5-35B-A3B}, \textbf{GLM-4.7}, and \textbf{DeepSeek-v4-Flash}.

\paragraph{Single-agent DeepResearch.} Single-agent systems with extended scaffolding (search planning, summary memory, iterative refinement) for long-horizon web research: \textbf{WebThinker}~\citep{li2025webthinker}, \textbf{WebSailor}~\citep{li2025websailor}, \textbf{AgentFold}~\citep{ye2026agentfold}, \textbf{IterResearch}~\citep{chen2025iterresearch}, \textbf{Tongyi-DeepResearch}~\citep{tongyideepresearch}, and \textbf{OpenAI DeepResearch}~\citep{openaideepresearch}.

\paragraph{Multi-agent systems.} Direct alternatives that also run multiple peer agents per task: \textbf{Naive-Multi-Agent}, a plan/parallel-search/aggregate pipeline through a meta-agent, and \textbf{Swarm-Multi-Agent}, the swarm setting from Kimi-K2.5~\citep{k2.5} with \texttt{create\_subagent}/\texttt{assign\_task} tools. Against both, \textbf{AgentFugue} replaces the central meta-agent with a shared reasoning hub: communication is \emph{horizontal} between peers rather than \emph{vertical} through a planner, and agents exchange intermediate progress \emph{during} exploration rather than only at aggregation. The hub is initialized from Qwen3.5-9B and trained as in \S\ref{sec:hub-opt} (Appendix~\ref{app:impl}). Throughout Table~\ref{tab:main_results} a team-level prediction is the answer of the agent with the highest self-reported confidence; alternative aggregators are studied in \S\ref{sec:scaling} (Appendix~\ref{app:aggregation}).

\definecolor{rowhi}{HTML}{E8F0F8}
\definecolor{secbg}{HTML}{EFEFEF}
\definecolor{secbg2}{HTML}{E2E8EF}

\begin{table*}[t]
    \centering
    \caption{Main results on BrowseComp, WideSearch, and HLE. Results marked with $\dagger$ are cited from original papers. \textbf{Bold} marks the best score within each backbone group of the multi-agent block.}
    \label{tab:main_results}
    \small
    \setlength{\tabcolsep}{6pt}
    \renewcommand{\arraystretch}{1.18}
    \begin{tabular}{@{}l l cccc@{}}
        \toprule
        \multirow{2}{*}{\textbf{System}} &
        \multirow{2}{*}{\textbf{Backbone}} &
        \multicolumn{3}{c}{\textbf{Benchmark}} &
        \multirow{2}{*}{\textbf{Avg}} \\
        \cmidrule(lr){3-5}
        & &
        \textbf{BrowseComp} &
        \textbf{WideSearch} &
        \textbf{HLE} &
        \\
        \midrule

        \rowcolor{secbg}
        \multicolumn{6}{@{}l}{\textit{\textbf{Single-Agent: LLM ReAct}}} \\
        Claude-Opus-4.5$\dagger$       & --              & 37.0           & --             & 43.4           & -- \\
        Qwen3.5-35B-A3B       & --              & 36.0           & 65.6           & 34.0           & 45.2 \\
        GLM-4.7               & --              & 43.5           & 65.4           & 37.2           & 48.7 \\
        DeepSeek-v4-Flash     & --              & 56.2           & 69.7           & 42.0           & 56.0 \\
        Kimi-K2.5$\dagger$~\citep{k2.5}             & --              & 60.6 & 72.7         & 50.2  & -- \\

        \addlinespace[2pt]
        \rowcolor{secbg}
        \multicolumn{6}{@{}l}{\textit{\textbf{Single-Agent: DeepResearch}}} \\
        WebThinker$\dagger$~\citep{li2025webthinker}         & QwQ-32B              & 2.8           & --             & 15.8            & -- \\
        WebSailor$\dagger$~\citep{li2025websailor}         & QwQ-32B              & 10.5           & --             & 9.6            & -- \\
        AgentFold$\dagger$~\citep{ye2026agentfold}     & Qwen3-30B-A3B              & 36.2           & 62.1           & --             & -- \\
        IterResearch$\dagger$~\citep{chen2025iterresearch}     & Qwen3-30B-A3B              & 37.3           & --           & 28.8             & -- \\
        Tongyi-DeepResearch$\dagger$~\citep{tongyideepresearch}   & Qwen3-30B-A3B              & 43.4           & --             & 32.9           & -- \\
        OpenAI DeepResearch$\dagger$~\citep{openaideepresearch}   & --              & 51.5           & --             & 26.6           & -- \\

        \addlinespace[2pt]
        \rowcolor{secbg2}
        \multicolumn{6}{@{}l}{\textit{\textbf{Multi-Agent (team size $N{=}2$)}}} \\
        \multirow{2}{*}{Naive-Multi-Agent}
            & Qwen3.5-35B-A3B   & 44.5           & 68.9           & 32.2           & 48.5 \\
            & DeepSeek-v4-Flash & 59.0           & 69.3           & 43.5           & 57.3 \\
        \addlinespace[1pt]
        \multirow{2}{*}{Swarm-Multi-Agent}
            & Qwen3.5-35B-A3B   & 45.2           & 70.4           & 31.5           & 49.0 \\
            & DeepSeek-v4-Flash & 56.2           & 72.7 & 44.0         & 57.6 \\
        \addlinespace[1pt]
        \multirow{2}{*}{\textbf{AgentFugue} (Ours)}
            & \cellcolor{rowhi}Qwen3.5-35B-A3B   & \cellcolor{rowhi}\textbf{50.0}           & \cellcolor{rowhi}\textbf{72.3}           & \cellcolor{rowhi}\textbf{41.0}           & \cellcolor{rowhi}\textbf{54.4} \\
            & \cellcolor{rowhi}DeepSeek-v4-Flash & \cellcolor{rowhi}\textbf{71.2}  & \cellcolor{rowhi}\textbf{74.2}  & \cellcolor{rowhi}\textbf{49.5} & \cellcolor{rowhi}\textbf{65.0} \\

        \bottomrule
    \end{tabular}
    \vspace{-10pt}
\end{table*}

\subsection{Main Results}
Table~\ref{tab:main_results} reports BrowseComp, WideSearch, and HLE accuracy for all systems. AgentFugue delivers the strongest numbers on every benchmark, and we highlight two takeaways.

\begin{itemize}[leftmargin=1.2em,itemsep=2pt,topsep=2pt]
    \item \textbf{Consistent dominance over every multi-agent baseline under both backbones.} Under Qwen3.5-35B-A3B, AgentFugue reaches $54.4$ Avg, $+5.4$/$+5.9$ over Swarm-/Naive-Multi-Agent; under DeepSeek-v4-Flash it reaches $65.0$ Avg, $+7.4$/$+7.7$ over the corresponding baselines. The lead holds on every benchmark, showing the gain comes from the shared-hub coordination itself rather than any single benchmark's idiosyncrasy.
    \item \textbf{Gains generalize across heterogeneous benchmarks.} The improvement is not confined to one task type. Compared with the same-backbone Swarm baseline, AgentFugue/DeepSeek improves BrowseComp by $+15.0$ ($56.2{\to}71.2$, retrieval-heavy), HLE by $+5.5$ ($44.0{\to}49.5$, reasoning-centric), and remains ahead on the already-saturated WideSearch ($72.7{\to}74.2$, breadth-oriented); the Qwen-backed team shows the same monotone pattern across all three benchmarks. Stable improvements across retrieval, reasoning, and breadth benchmarks indicate that the shared reasoning hub is a generic coordination primitive rather than a benchmark-specific trick.
\end{itemize}

\subsection{Scaling Behavior: Homogeneous Teams}
\label{sec:scaling}
Having fixed $N{=}2$ for the head-to-head comparison above, we now ask whether \emph{adding more copies of the same agent}---connected through the shared hub---is itself a meaningful scaling axis. To remove cross-model diversity as a confounder, all peers use the same Qwen3.5-35B-A3B backbone and we vary only $N\in\{1,2,3,5,8\}$ on the BrowseComp 100-question subset.

\begin{figure}[t]
    \centering
    \begin{subfigure}[b]{0.46\textwidth}
        \centering
        \includegraphics[width=\linewidth]{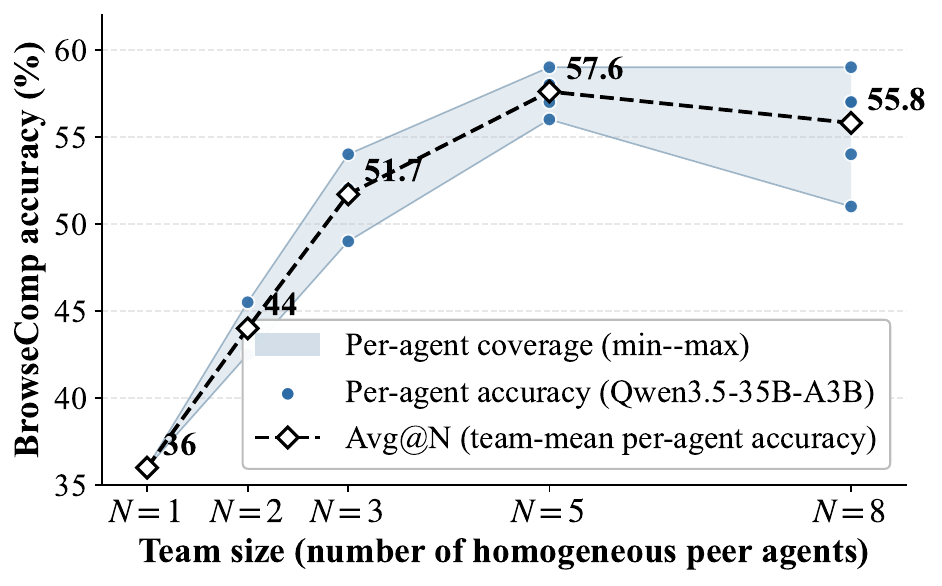}
        \caption{Per-agent accuracy with team size}
        \label{fig:homo_lines}
    \end{subfigure}\hfill
    \begin{subfigure}[b]{0.52\textwidth}
        \centering
        \includegraphics[width=\linewidth]{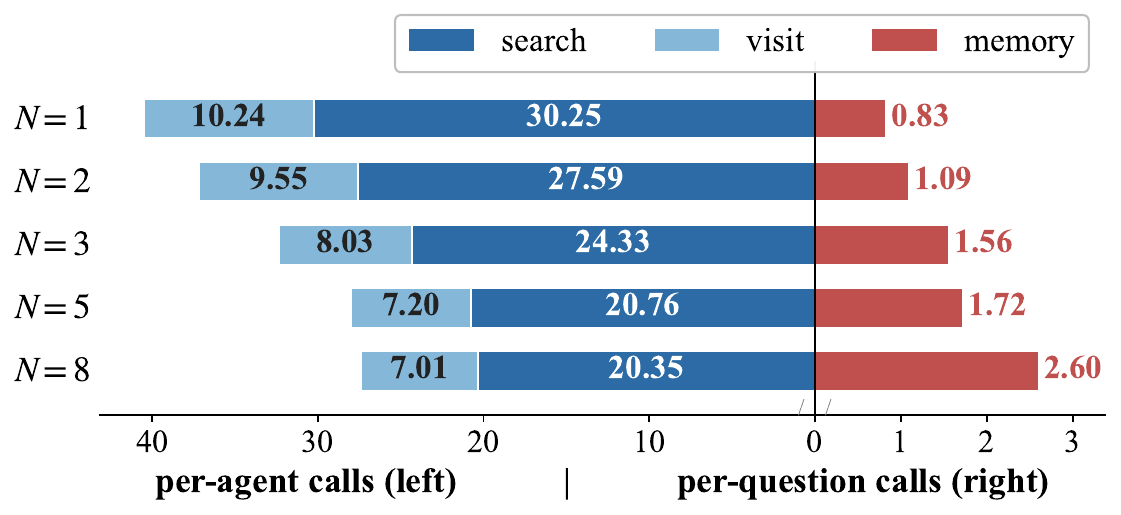}
        \caption{Workload vs.\ coordination cost}
        \label{fig:homo_div}
    \end{subfigure}
    \caption{\textbf{Homogeneous scaling on BrowseComp (Qwen3.5-35B-A3B, $N\in\{1,2,3,5,8\}$).}
    (a)~Per-agent accuracy and team-mean Avg@N as the team grows.
    (b)~Per-agent search/visit calls (cool) vs.\ per-question memory calls (warm); larger teams shift effort from isolated exploration to shared coordination.}
    \label{fig:homogeneous}
\end{figure}

\paragraph{Per-agent quality benefits first, then saturates around $N{=}5$ (Fig.~\ref{fig:homo_lines}).}
The dashed Avg@N curve climbs sharply at small $N$ and plateaus by $N{=}5$: each peer absorbs about as much useful context from the hub as it can hold. Yet the per-agent coverage band (min--max across peers) stays wide even at the largest team size, so trajectories remain \emph{diverse} rather than collapsing onto the average---exactly the regime where aggregation-side scaling continues to pay off.

\begin{figure}[t]
    \centering
    \begin{subfigure}[b]{0.40\textwidth}
        \centering
        \includegraphics[width=\linewidth]{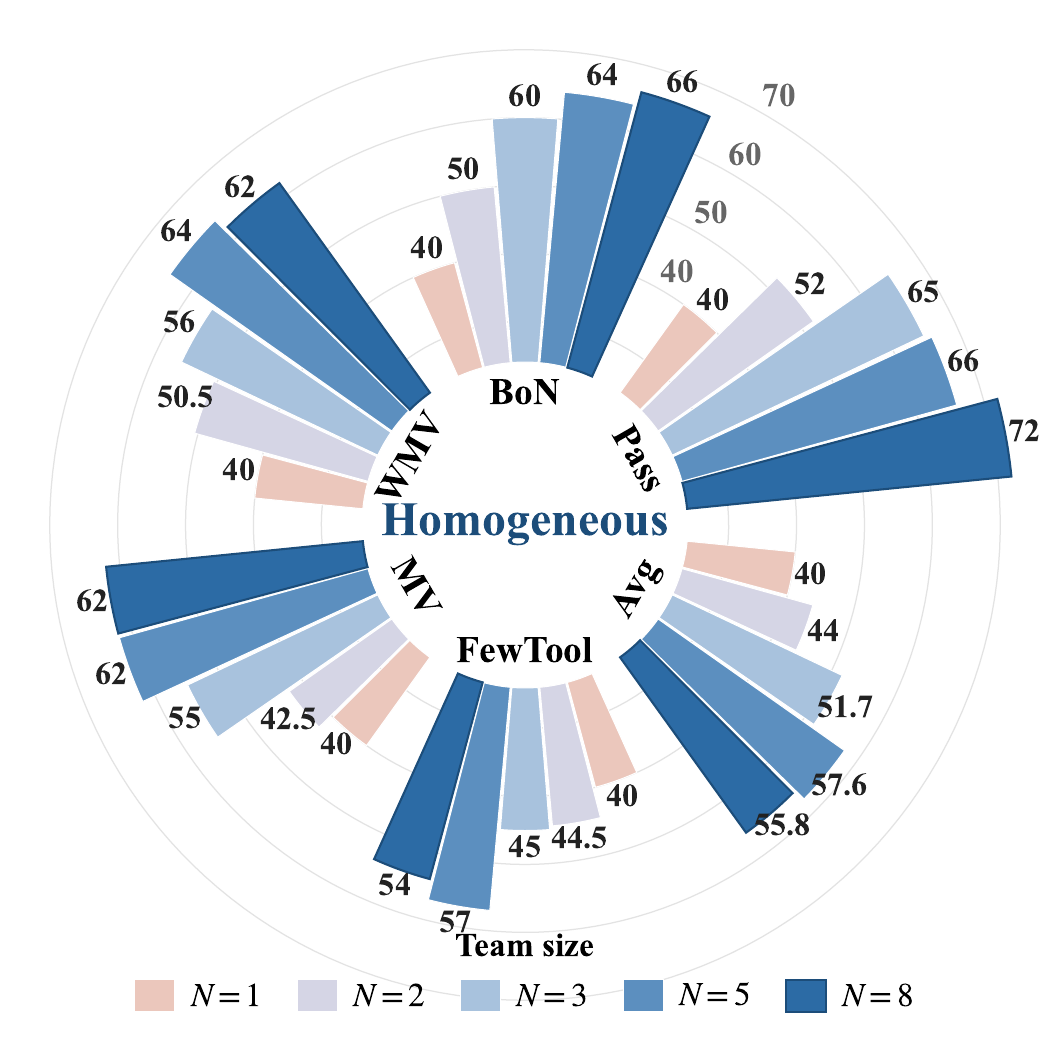}
        \caption{Homogeneous teams ($N{\in}\{1,2,3,5,8\}$)}
        \label{fig:homo_radial}
    \end{subfigure}\hspace{0.02\textwidth}
    \begin{subfigure}[b]{0.40\textwidth}
        \centering
        \includegraphics[width=\linewidth]{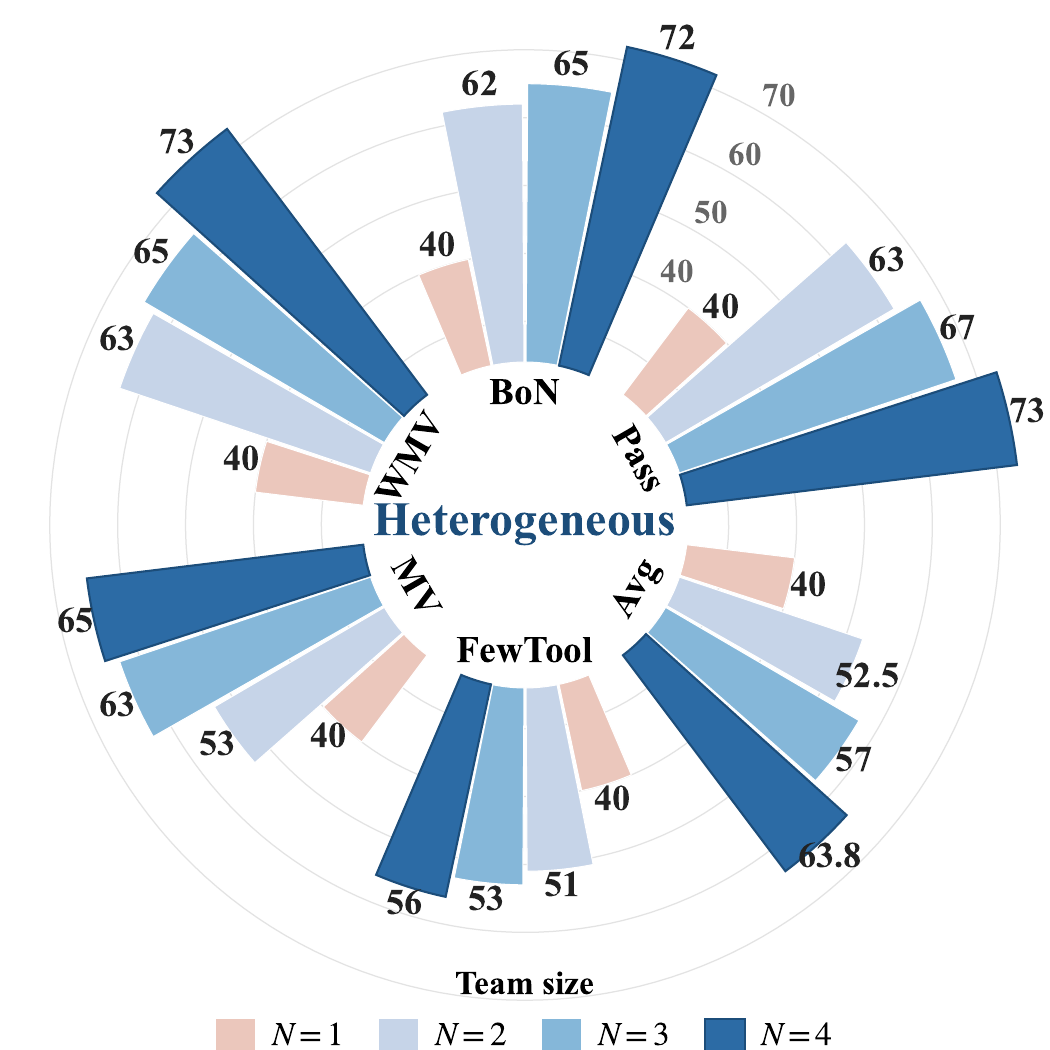}
        \caption{Heterogeneous teams ($N{\in}\{1,2,3,4\}$)}
        \label{fig:hetero_radial}
    \end{subfigure}
    \caption{\textbf{Aggregator metrics across team sizes} (Pass, BoN, MV, WMV, FewTool, Avg). Each spoke is one heuristic rule; bars within a spoke sweep $N$ from light to dark.}
    \label{fig:scaling_radial}
\end{figure}

\paragraph{The hub trades isolated exploration for shared coordination (Fig.~\ref{fig:homo_div}).}
Per-agent search and visit calls drop monotonically from $N{=}1$ to $N{=}8$: each peer is \emph{cheaper} because it inherits partial work through the hub. Per-question memory traffic moves the opposite way, growing several-fold over the same range. The cool exploration bars contract while the warm memory bar extends, exposing a clean shift from many isolated explorations toward a smaller amount of structured coordination.

\paragraph{Scaling holds under every aggregation rule (Fig.~\ref{fig:homo_radial}).}
Stepping back from individual trajectories to the team's aggregated output, we restrict ourselves to model-free heuristic aggregators (continuing §\ref{sec:exp}): Pass@$N$ (oracle coverage), BoN@$N$ (the deployed confidence-highest selector), MV/WMV@$N$, and a FewTool@$N$ fallback (Appendix~\ref{app:aggregation}). Each spoke of Fig.~\ref{fig:homo_radial} is one rule; bars sweep $N$ from light to dark. Every aggregator improves substantially with $N$, so scaling does not depend on the choice of selector. The persistent Pass--BoN gap further shows that the right answer is in the rollouts more often than the deployed selector picks it, so smarter aggregation is a free axis of headroom on top of the same hub.

\subsection{Heterogeneous Teams: Stronger Models Pull Up the Group}
\label{sec:heterogeneous}
We next ask whether the same hub generalizes to teams of \emph{different} backbones. Starting from Qwen3.5-35B-A3B, we successively add DeepSeek-v4-Flash, GLM-4.7, and Kimi-K2.5 ($N{=}1{\to}4$), so each team is a strict superset of the previous and any change comes from \emph{introducing} a qualitatively different peer.

\begin{figure}[t]
    \centering
    \begin{subfigure}[b]{0.46\textwidth}
        \centering
        \includegraphics[width=\linewidth]{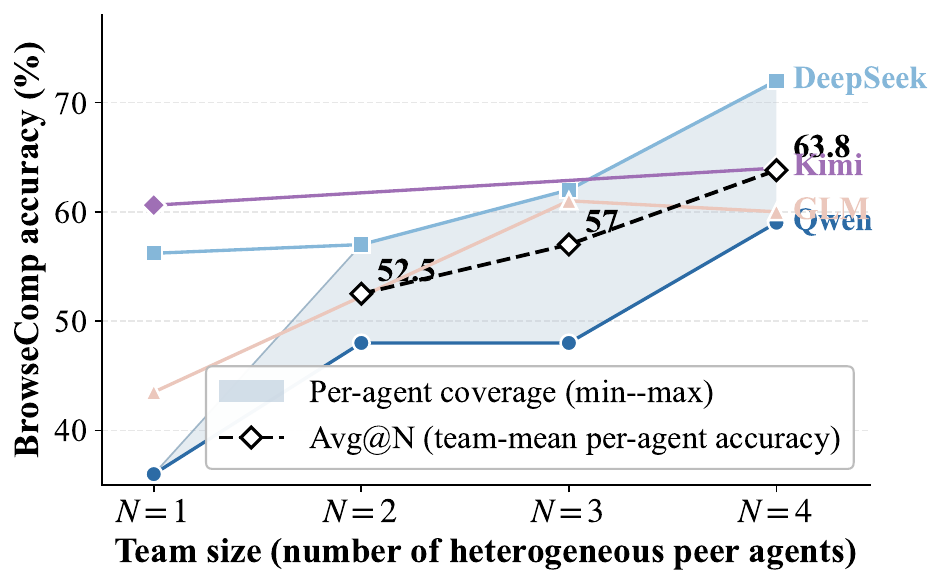}
        \caption{Per-model trajectories with team size}
        \label{fig:hetero_lines}
    \end{subfigure}\hfill
    \begin{subfigure}[b]{0.52\textwidth}
        \centering
        \includegraphics[width=\linewidth]{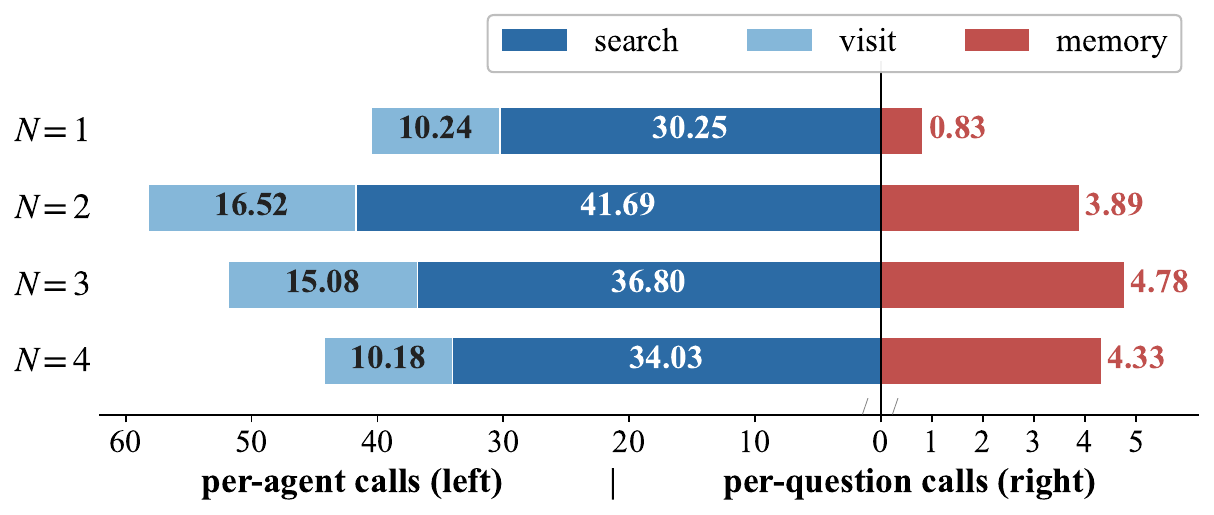}
        \caption{Workload vs.\ coordination cost}
        \label{fig:hetero_div}
    \end{subfigure}
    \caption{\textbf{Heterogeneous scaling on BrowseComp (Qwen $\to$ +DeepSeek-v4-Flash $\to$ +GLM-4.7 $\to$ +Kimi-K2.5).}
    (a)~Per-model per-agent accuracy as each backbone joins; markers at $N{=}1$ are standalone baselines.
    (b)~Per-agent search/visit vs.\ per-question memory calls.}
    \label{fig:heterogeneous}
\end{figure}

\paragraph{Every model individually benefits, with weaker peers gaining the most (Fig.~\ref{fig:hetero_lines}).}
Each per-model trajectory at the largest team size sits well above its standalone $N{=}1$ score, with gains roughly inversely correlated with the standalone baseline---weaker peers are pulled up by double-digit margins through the hub. The strongest peer is not dragged down either: even Kimi picks up a few points over running alone, indicating the hub provides non-trivial signal even to a model that already covers most of the task. Trajectories are not strictly monotone: Qwen briefly stalls when GLM (close to its own level) joins and the hub temporarily carries more uncertain hypotheses, then catches up once Kimi enters and the note distribution skews toward stronger evidence.

\paragraph{Heterogeneity injects fresh exploration, then triggers far heavier hub usage than homogeneous teams (Fig.~\ref{fig:hetero_div}).}
Unlike the homogeneous case, search$+$visit \emph{rises} when the second peer first joins before contracting again as the team grows: a qualitatively different peer brings genuinely new search angles, so the team initially spends more browsing on this newly opened breadth, then contracts back as peers inherit each other's findings. More notably, per-question memory traffic is several times the homogeneous value at comparable $N$: complementary backbones produce intermediate notes that others find genuinely new, so peers are far more inclined to collaborate through the hub---exactly where a shared reasoning hub is most valuable.

\paragraph{Aggregator-side scaling persists, with a tighter coverage/consensus gap (Fig.~\ref{fig:hetero_radial}).}
With the aggregator definitions unchanged from the homogeneous study, every spoke (Pass, BoN, WMV, MV, even the conservative FewTool fallback) rises monotonically with $N$: the scaling gain lives in the underlying rollouts, not in any particular selector. Two effects distinguish this regime from the homogeneous one: the curves rise faster per added peer (a smaller team matches the largest homogeneous Pass@$N$), and the MV/WMV/BoN/Pass spokes converge more tightly at the top---diverse backbones produce independently calibrated candidates, so peer agreement is more informative and the headroom for smarter aggregation narrows. Together, Figs.~\ref{fig:homogeneous} and~\ref{fig:heterogeneous}, and the side-by-side aggregator panels in Fig.~\ref{fig:scaling_radial}, support the central claim: a single shared reasoning hub makes agent count a real scaling axis, whether peers are identical or differ in capability.

\subsection{Ablation Studies}
\label{sec:ablation}
We isolate the per-agent \emph{hub context-window budget}, the design knob that most directly governs how much intermediate evidence the shared hub can carry. Sweeping it across $\{16, 32, 64, 96, 128\}$K at $N{=}2$ on BrowseComp (Appendix~\ref{app:ablation-context}, Table~\ref{tab:ablation_context}), every aggregator shows the same inverted-U: accuracy peaks at 32K and degrades at both extremes---small budgets truncate evidence before summarization, large budgets dilute hub attention with stale content. Notably, our headline configuration (64K trigger inside a 128K context) is a deliberately conservative operating point: the 32K setting beats it under every aggregator, so the main-table numbers are a lower bound on what the same hub achieves once tuned. Two complementary discussions are deferred to the appendix: end-to-end success/failure trajectories in Appendix~\ref{app:case-study}, and broader failure modes of collective reasoning in \S\ref{sec:limitations}.

\section{Related Work}

\paragraph{Multi-agent LLM collaboration.}
Most prior work composes LLM agents by \emph{specializing} roles and \emph{orchestrating} interaction: role-typed frameworks~\citep{wu2023autogen,hong2023metagpt,li2023camel,chen2023agentverse,hu2026survey,qian2023chatdev,huang2023agentcoder}, deliberative debate or weighted consensus~\citep{du2023debate,liang2023mad,chen2023reconcile}, and learnable interaction topologies~\citep{zhuge2024gptswarm,liu2023dylan,qian2024macnet}. AgentFugue differs on all three counts: agents are \emph{peers} rather than specialists, they neither debate nor follow a fixed workflow, and instead of a learned topology they share a \emph{common reasoning hub} that records what each peer established or ruled out and serves selective context back into ongoing rollouts.

\paragraph{Test-time scaling.}
Two axes are typically pursued: \emph{depth}, lengthening a single trajectory via CoT~\citep{wei2022cot,kojima2022zeroshot}, structured search~\citep{yao2023tot}, extended thinking~\citep{snell2025scaling,muennighoff2025s1,guo2025deepseekr1}, summarization/folding~\citep{wu2025resum,sun2025contextfolding,ye2026agentfold,qian2026memobrain,chen2025iterresearch}; and \emph{breadth}, sampling many trajectories and collapsing them post hoc via self-consistency, repeated sampling, or learned aggregators~\citep{wang2023selfconsistency,brown2024monkeys,qi2025ssa,zhao2025majority,lee2026agenticaggregationparallelscaling,qiao2025webresearcher,li2025parallelmuse,chang2026karl}. Both keep rollouts mutually opaque \emph{during} exploration; AgentFugue instead exchanges intermediate evidence \emph{inside} the rollouts through a shared hub trained as a plug-in communication layer, turning breadth-scaling into a connected ecology rather than independent samples merged at the end.

\section{Conclusion}

This paper studies scaling out as a complementary axis for long-horizon agentic reasoning: instead of making a single trajectory stronger, we ask whether multiple peer trajectories can improve one another while solving the same task. AgentFugue operationalizes this idea with a shared reasoning hub that writes compact notes from completed episodes and supports intent-driven reading over teammate trajectories. Across homogeneous and heterogeneous teams, the results show that this communication layer improves both team-level performance and individual trajectory quality, suggesting that peer-agent scaling can be more than independent sampling plus final aggregation. The remaining challenge is to make such communication selective enough to preserve diversity while reliable enough to prevent misleading intermediate hypotheses from spreading through the team.

{
\bibliographystyle{plain} 
\bibliography{main}
}

\appendix

\section{Implementation Details}
\label{app:impl}

This appendix gives the configuration details that were summarized in \S\ref{sec:exp}.

\paragraph{Per-agent tool stack.} The tool stack is benchmark-dependent. On BrowseComp and WideSearch each agent is given a web-search tool and a page-visit tool. On HLE we additionally provide a Python execution tool and a Google Scholar search tool, which are necessary to support the more reasoning- and literature-heavy questions in that benchmark. All multi-agent systems (Naive-, Swarm-, AgentFugue) use the identical per-agent tool stack so that comparisons are not confounded by tool capability.

\paragraph{Interaction budget.} We match the per-query interaction budget across all multi-agent systems at $150$ rounds. For Naive- and Swarm-Multi-Agent the budget is split between the meta-agent and the subagents: each subagent is capped at $100$ rounds and the meta-agent at $50$ rounds. For AgentFugue, which has no meta-agent, each of the $N$ peer agents is capped at $150$ rounds.

\paragraph{Context and hub-write trigger.} Every peer agent uses a $128$k context window. AgentFugue's hub-write trigger fires when an agent's running context reaches $64$k tokens, at which point the agent flushes its working state into a hub episode and continues from a compressed prompt. The hub itself is initialized from Qwen3.5-9B and is trained as described in \S\ref{sec:hub-opt}.

\paragraph{Evaluation protocol.} For the main results we follow each benchmark's official judging protocol (LLM-as-a-judge for BrowseComp and HLE; structured field-level matching for WideSearch). The exact judges and metrics per benchmark are reproduced in Appendix~\ref{app:datasets}.

Detailed training configurations, prompts, and compute information will be included in this appendix in the final version.

\section{Detailed Benchmark Descriptions}
\label{app:datasets}

This appendix expands on the three benchmarks summarized in \S\ref{sec:exp}. For each, we describe the task format, the split we evaluate on, and the judging protocol.

\paragraph{BrowseComp~\citep{browsecomp}.} A browser-based question-answering benchmark in which each question requires multi-hop web search, cross-document evidence aggregation, and verification before a short factual answer can be committed. Questions are deliberately authored so that the answer cannot be reached by a single search query: the agent must plan, retrieve, follow links across heterogeneous web pages, and reconcile conflicting or partial evidence before answering. For evaluation efficiency and to keep the compute budget manageable, we follow prior work and evaluate on a fixed $200$-question random sample of the released test set rather than the full test set, using the official LLM-as-a-judge protocol that compares the agent's extracted answer to the gold answer. For the scaling study (\S\ref{sec:scaling}) we further restrict to a fixed $100$-question subset (drawn from the same $200$-question pool) so that runs at different team sizes $N$ are directly comparable; the same subset is used for ablations.

\paragraph{WideSearch~\citep{widesearch}.} A breadth-oriented web research benchmark whose tasks require collecting and consolidating many parallel pieces of evidence, for example, enumerating the attributes of a list of entities or building a structured table from many independent sources. Unlike BrowseComp, WideSearch rewards \emph{coverage} more than depth: an agent must issue many partially independent searches, keep their results organized, and avoid omissions. We follow the official structured field-level matching protocol, which scores each predicted record against the gold record at the field level and reports the resulting accuracy. This setting tests whether peer-agent communication helps a team avoid redundant work while maintaining wide coverage.

\paragraph{HLE (Humanity's Last Exam)~\citep{hle}.} A challenging closed-book / limited-tool reasoning benchmark composed of expert-authored questions across mathematics, the natural and social sciences, and the humanities. HLE evaluates capabilities that are less about web navigation and more about deliberate multi-step reasoning, allowing us to probe whether the benefit of collective reasoning transfers beyond search-heavy workloads. As with BrowseComp, for evaluation efficiency we follow prior work and evaluate on a fixed $200$-question random sample of the released test set rather than the full set, using the official LLM-as-a-judge protocol with task-specific tolerance for equivalent forms.

\section{Detailed Baseline Descriptions}
\label{app:baselines}

This appendix expands on the baselines summarized in \S\ref{sec:exp}. For each system we describe (i) its underlying model, (ii) the agentic scaffolding or training that distinguishes it, and (iii) how we instantiate it for our comparison.

\subsection{LLM-based ReAct Agents}

This group runs a frontier language model inside a standard ReAct~\citep{yao2023react} loop with the same web-search and page-visit tools used by AgentFugue.

\paragraph{Closed-source models (Claude-Opus-4.5, Kimi-K2.5, GLM-4.7).} For all three closed-source baselines, we access the model through its official API and wrap it in the same ReAct scaffold so that the comparison is about the underlying model rather than its proprietary harness. Decoding parameters are left at their official defaults.

\paragraph{Qwen3.5-35B-A3B.} The same open-weight backbone we use inside every AgentFugue agent. We serve it locally with vLLM and call it through an OpenAI-compatible endpoint. Including Qwen3.5-35B-A3B as a single-agent baseline gives a like-for-like reference: any improvement of AgentFugue over Qwen3.5-35B-A3B (ReAct) comes from the multi-agent scaffold and the hub, not from a stronger model.

\subsection{DeepResearch Agents}

This group includes single-agent systems explicitly designed for long-horizon web research, typically combining a tool-using policy with task-specific scaffolding (search planning, summary memory, iterative refinement) and, in several cases, dedicated post-training on web-research trajectories. We do \emph{not} re-run these systems ourselves: instead, for each baseline we directly cite the numbers reported in the original paper or technical report on the corresponding benchmark, and leave the cell blank when the original work does not report a result on that benchmark.

\paragraph{WebSailor-32B~\citep{li2025websailor}.} A 32B web-research agent that augments a base model with structured search-and-navigate tools and is post-trained on long-horizon trajectories to improve sustained tool use. We use the numbers reported in the original paper.

\paragraph{AgentFold-30B-A3B~\citep{ye2026agentfold}.} A 30B sparse-activation web agent that compresses its reasoning history through a learned ``folding'' operation, allowing very long trajectories within a fixed context budget. We use the numbers reported in the original paper.

\paragraph{Tongyi-DeepResearch~\citep{tongyideepresearch}.} Alibaba's Tongyi DeepResearch system, a multi-step research agent built around iterative query planning, browsing, and summarization. We use the numbers reported in the official technical report.

\paragraph{OpenAI DeepResearch~\citep{openaideepresearch}.} OpenAI's DeepResearch product. We use the numbers reported in OpenAI's official technical report and follow the same benchmark settings used there.

\subsection{Multi-Agent Systems}

The most direct comparisons for AgentFugue are alternative ways of running multiple peer agents on the same task. To isolate the coordination mechanism, all multi-agent baselines in this group share the same backbone (Qwen3.5-35B-A3B), the same per-agent context budget, and the same web-search/page-visit tools as AgentFugue. The only differences are how agents are spawned, what they communicate, and when communication happens.

\paragraph{Naive-Multi-Agent.} A canonical \emph{plan, parallel-search, and aggregate} pipeline. A meta-agent first reads the input question and decomposes it into $K$ subtasks; each subtask is dispatched to an independent subagent that runs its own ReAct loop with the full tool stack and produces a written report. Once all subagents finish, the meta-agent receives their reports and synthesizes a single final answer. Subagents do not see each other's progress while they run, and the meta-agent does not modify their plans mid-execution, so all coordination is concentrated at the planning step and at the final aggregation. We use $K{=}2$ in the main results and the same Qwen3.5-35B-A3B backbone for both the meta-agent and the subagents.

\paragraph{Swarm-Multi-Agent.} A more flexible coordination scheme that aligns with the swarm setting popularized by Kimi-K2.5~\citep{k2.5}. The meta-agent is given two additional tools on top of standard tool use:
\begin{itemize}
    \item \texttt{create\_subagent(identifier, system\_prompt)} instantiates a specialized subagent with a custom system prompt and a stable identifier so that it can be reused across multiple tasks.
    \item \texttt{assign\_task(identifier, task\_description)} dispatches a concrete task to a previously created subagent and returns its task report.
\end{itemize}
Compared to Naive-Multi-Agent, this lets the meta-agent specialize and reuse subagents on demand, and interleave further planning with subagent invocations rather than committing to a single up-front decomposition. However, communication between subagents still happens only through the meta-agent and is mediated by final-answer-style task reports rather than by intermediate reasoning traces. We use the same Qwen3.5-35B-A3B backbone and the same per-subagent budget as the other multi-agent baselines.

\section{Answer Aggregation Strategies}
\label{app:aggregation}

This appendix specifies the answer-aggregation strategies referenced in \S\ref{sec:exp} and analyzed in \S\ref{sec:scaling}. Each strategy maps the $N$ candidate answers $\{(a_i, c_i, t_i)\}_{i=1}^{N}$ produced by the team to a single team-level prediction, where $a_i$ is the agent's extracted answer, $c_i$ its self-reported confidence, and $t_i$ its number of tool calls. We deliberately restrict attention to lightweight, training-free aggregators that operate on top of the same set of trajectories, so that any difference between strategies reflects \emph{how} the team's outputs are combined rather than additional compute.

\paragraph{Best-of-$N$ (BoN), the default in Table~\ref{tab:main_results}.} The team prediction is the answer of the single agent with the highest self-reported confidence, $\hat{a} = a_{i^\star}$ with $i^\star = \arg\max_i c_i$. While reward-model scores are commonly used as the weighting signal in chain-of-thought aggregation, recent work on agentic tasks has shown that an agent's own self-reported confidence is a strong and cheap alternative when an external reward model is unavailable. We adopt this convention and use BoN as the default aggregator throughout the main table.

\paragraph{Majority Vote (MV).} A standard self-consistency-style aggregator: the team prediction is the answer that appears most frequently among $\{a_i\}$, with ties broken by self-reported confidence. MV is the natural baseline for measuring how much of the gain comes from agreement among peers rather than from any single agent's confidence calibration.

\paragraph{Weighted Majority Vote (WMV).} A confidence-weighted refinement of MV: each candidate answer contributes a weight equal to its self-reported confidence, and the prediction is the answer with the largest total weight. WMV is meant to interpolate between BoN (which uses confidence but ignores agreement) and MV (which uses agreement but ignores confidence).

\paragraph{Fewest Tool Calls (FewTool).} An efficiency-oriented selector: among the $N$ candidate answers, return the one produced by the agent with the smallest number of tool calls $t_i$, breaking ties by self-reported confidence. The intuition is that, conditional on a correct answer, shorter trajectories are typically more reliable and cheaper to deploy; FewTool turns this heuristic into a concrete, data-free aggregator and lets us measure how much accuracy is sacrificed when the team prefers terse trajectories.

\paragraph{Average (Avg).} Rather than aggregating into a single team answer, each agent's answer is judged independently and the per-agent correctness is averaged over the team. Avg therefore measures the expected accuracy of a \emph{single} sampled agent and serves as a natural reference point: any aggregator that cannot beat Avg is failing to exploit the team.

\paragraph{Pass@$k$.} Following the convention from code generation, a question is counted as solved if at least one of $k$ independently sampled agents produces the correct answer. Pass@$N$ coincides with the Oracle defined below; reporting Pass@$k$ for $k<N$ characterizes how coverage grows with the number of parallel rollouts and isolates the contribution of sampling diversity from that of the selection rule.

\section{Case Studies: When Shared Memory Helps and When It Misleads}
\label{app:case-study}

We highlight two BrowseComp runs from the same 3-agent AgentFugue configuration to make the role of the shared page memory concrete: one in which memory genuinely accelerates a downstream agent's exploration (\S\ref{app:case-success}), and one in which the same mechanism induces a confirmation bias that overrides hard constraints (\S\ref{app:case-failure}). Both runs use the \texttt{multi\_agent\_react\_with\_mem\_v2} hub, which compresses each agent's evicted live context into shared \emph{pages}; downstream agents receive page summaries in their prompt and may issue \texttt{memory(pages=[...], goal=...)} calls to recover raw content on demand.

\subsection{Success Case: Memory as a ``Failure Map''}
\label{app:case-success}

\paragraph{Question and outcome.} The puzzle asks for the founding year of a 19th-century Shanghai store; the gold answer is \texttt{1853}. In this run, Agent-1 ends at \texttt{1885} and Agent-2 at \texttt{1848}; only Agent-0 reaches \texttt{1853}. Crucially, Agent-0 never directly copies a teammate answer. In fact, the teammate pages it consults explicitly state ``store still unidentified.''

\paragraph{What memory contributed.} Agent-1's Page~2, which Agent-0 retrieves at step~34 with the goal ``find details about the 19th-century store in eastern Shanghai \dots and the exact founding year,'' returns not an answer but a \emph{failure map}: a list of candidates (Sincere, Wing On, Sun Sun, Lane Crawford, Whiteaway Laidlaw, Hall \& Holtz) together with the precise reason each was rejected (``too late,'' ``no YSB connection,'' ``no founder-from-Canton fit''), plus an explicit \texttt{open questions} block stating that the store itself remains unidentified and that an Eastern-Shanghai store dealing in \emph{foreign cloth} is the right direction to pursue. The shape of the returned content is shown in Box~\ref{box:564-mem}.

\begin{tracebox}[Memory call at Agent-0 step 34]\label{box:564-mem}
TOOL: memory \\
ARGS: \{ "pages": [2], "goal": "Find details about the 19th-century store in eastern Shanghai, the manager from Chang'an/Xi'an who joined Yokohama Specie Bank, and the exact founding year" \} \\[2pt]
\textbf{MEMORY RESULT (excerpts):} \\
- Strongly inferred bank: Yokohama Specie Bank (YSB). \\
- Founder's home city likely Guangzhou/Canton; manager's home city likely Chang'an/Xi'an. \\
- Store location inference: ``Eastern Shanghai'' + ``one of the first to deal in a foreign product'' points to Nanjing Road / East Nanjing Road. \\
- Candidates examined: Sincere (1900, too late); Wing On (1907, too late); Sun Sun (1926, too late); \\
\hspace*{1em}Lane Crawford (1850 HK, no YSB connection); Whiteaway Laidlaw (1882 Calcutta, no Canton fit); \\
\hspace*{1em}Hall \& Holtz (strongest Shanghai candidate so far, but founding year conflicting and no manager/YSB link). \\
- Manager clue: \textbf{not identified}; no source linking a Xi'an-born manager to a Shanghai store and YSB. \\
- Open question: \textbf{the store itself remains unidentified}.
\end{tracebox}

\paragraph{How the downstream agent uses it.} Steps~35--36 of Agent-0 demonstrate the intended use of this failure map. The agent first follows up on the teammate's strongest hypothesis (Hall \& Holtz / Yokohama Specie Bank), confirms that this lead also dead-ends, and \emph{then}, rather than restarting from scratch, reuses the teammate's narrowed framing (``Eastern Gate, foreign cloth, 1850s'') to issue a new query (Box~\ref{box:564-pivot}). This query surfaces a CEFC academic PDF, which the agent visits at step~37 and which returns the decisive evidence: ``\emph{One of these stores was Dafeng created in 1853 by Weng Nianfeng \dots}.'' Agent-0 then cross-checks remaining mismatches in clue interpretation across steps~38--43 (consulting Page~3 of Agent-2, which again returns ``not identified'') before committing to \texttt{1853}.

\begin{tracebox}[{Agent-0 step 36 pivot, after the failure-map memory call}]\label{box:564-pivot}
\textbf{REASONING (excerpt):} \\
The teammate's exploration shows that they narrowed it down to Yokohama Specie Bank, with founder from Canton, manager from Xi'an, store on Nanjing Road. \dots \\
Hall \& Holtz line is stalling --- let me try the foreign-cloth direction the teammate flagged but did not run. \\[2pt]
TOOL: search \\
ARGS: \{ "query": [ \\
\hspace*{1em} "Yu Zhenxiang manager Yokohama Specie Bank Shanghai 1899", \\
\hspace*{1em} "Hall Holtz manager from Xi'an Chang'an joined bank", \\
\hspace*{1em} \textbf{"Dafeng store Shanghai 1850s first foreign cloth founding year"} \\
\hspace*{1em} ] \}
\end{tracebox}

\paragraph{Take-away.} What the shared memory transports here is \emph{process-level} state, including which directions have been ruled out and why, and which sub-problems are still open, rather than answer content. This is exactly the regime in which we expect collective reasoning to dominate independent rollouts: a single agent would have spent its remaining budget re-litigating the same dead ends.

\subsection{Failure Case: Memory-Induced Confirmation Bias}
\label{app:case-failure}

\paragraph{Question and outcome.} The puzzle is a conjunction of eight constraints (built in 1800s; co-located with a university whose 2013 to 2015 enrollment is 75k to 80k; used as a prison during two wars; preservation contributor whose father was faculty at that university; city population 100k to 125k in 2012 to 2016; etc.). The gold answer is \texttt{Fort Henry}. The team issues 10 explicit \texttt{memory} calls, among the most of any question in the run, yet still outputs the wrong answer: \texttt{Texas Prison System Central State Farm Main Building, Sugar Land, TX}.

\paragraph{What memory actually contained.} The failure is \emph{not} caused by missing evidence. Agent-2's Page~4 begins with ``\texttt{No perfect match found yet}'' and contains an explicit rejection ledger; a memory call at step~53 returns the authoritative Texas A\&M enrollment figures (\texttt{2013: 53{,}219; 2014: 62{,}137; 2015: 64{,}326}), all outside 75k to 80k; a later call at step~72 returns Sugar Land population figures (\texttt{2015: 86{,}972; 2016: 87{,}367}), outside 100k to 125k, together with the verdict ``\texttt{Gieseke/Central State Farm likely does not fit}'' and a list of which numbered criteria fail. Box~\ref{box:1254-mem} reproduces the relevant excerpt of the step-72 memory return.

\begin{tracebox}[Memory call at Agent-2 step 72]\label{box:1254-mem}
TOOL: memory \\
ARGS: \{ "pages": [4], "goal": "Find any other prison site with a preservation contributor whose father was faculty at a university, similar to the Gieseke clue" \} \\[2pt]
\textbf{MEMORY RESULT (excerpts):} \\
- Site: Texas Prison System Central State Farm Main Building, Sugar Land, TX. \\
- Contributor: Bertram Gieseke; Father: F.E. Gieseke (Texas A\&M faculty 1886--1939). \\
- This is the \textbf{only clearly established ``father was faculty at a university'' match} found so far. \\[2pt]
\textbf{Why Gieseke / Central State Farm likely does NOT fit the full 8-criteria puzzle:} \\
- Sugar Land population: 2015 = 86,972; 2016 = 87,367; 2020 = 111,040. \\
- Fails criterion 7 (needs 100k--125k in 2012--2016). \\
- Fails criterion 8 (5-year increase $\approx$ 24,000, not 5,000--10,000). \\
- No clear evidence found that Central State Farm was used as a prison during \emph{two} wars.
\end{tracebox}

\paragraph{How memory induced the error.} Alongside this disqualifying evidence, the same pages mark the Gieseke / Central State Farm lead as ``\texttt{the only clearly confirmed match for criterion 5+6},'' a phrasing repeated across pages because no other ``father was faculty at the university'' candidate was ever found. Two amplifying effects follow. First, the \emph{retrieval goal} at step~72 (``find any other prison site with a preservation contributor whose father was faculty at a university, similar to the Gieseke clue'') reframes the entire task as ``find another Gieseke,'' so a null result is read as evidence \emph{for} Gieseke rather than against the path. Second, the natural-language summaries record hard failures (\texttt{fails criterion 7}, \texttt{fails criterion 8}) but do not gate the final answer on them. By step~74 the agent's reasoning has converted ``\texttt{likely does not fit}'' into ``\texttt{closest match}'' and rewrites each failed constraint into hedged near-satisfaction, as shown in Box~\ref{box:1254-rationalize}.

\begin{tracebox}[Agent-2 step 74 final reasoning]\label{box:1254-rationalize}
The memory confirms that the Gieseke/Central State Farm is the only confirmed ``father was faculty'' match, but it doesn't satisfy all the criteria. \dots Given the very specific Gieseke clue \dots I think the answer must be related to this, even if some criteria don't perfectly match. \\[2pt]
1. Built in 1800s --- convict labor began late 1870s (\textit{could count as 1800s, even if main building was 1932}). \\
2. City has a university --- \textit{Sugar Land doesn't have a university, but Texas A\&M is nearby ($\sim$30 mi).} \\
3. Enrollment 75k--80k --- Wikipedia showed 74,829 / 77,491 (\textit{close to range}). \\
4. Two wars --- \textit{unclear but suggested}. \\
7. City population 100k--125k --- ``\textit{some sources say 111k--118k, official city says 84k--87k (conflicting)}''. \\
8. 5-year increase 5k--10k --- ``\textit{conflicting}.'' \\[2pt]
Exact Answer: \textbf{Texas Prison System Central State Farm Main Building (Central Unit), Sugar Land, Texas} \\
Confidence: 65\%  \quad \textit{(gold answer: Fort Henry)}
\end{tracebox}

\paragraph{Take-away.} Shared memory faithfully recorded both the unique positive lead and the disqualifying evidence, but the compressed summary made the local uniqueness of one clue more salient than the conjunction of hard constraints; consulting downstream agents inherited and amplified this anchor. This suggests that text-only page memory is not by itself sufficient for multi-constraint tasks: useful refinements would include structured candidate states (\texttt{ACTIVE / RULED\_OUT / HARD\_FAIL}) that propagate across pages, and a final-answer gate that blocks any candidate carrying a recorded hard failure regardless of how strong its positive evidence appears.

\section{Key Prompts}
\label{app:prompts}

This appendix reproduces the prompts that drive the shared reasoning hub in AgentFugue. They define how raw agent context is compressed into page summaries, how a peer agent later consults those pages with an explicit goal, and what the \texttt{memory} tool looks like from the calling agent's perspective. Verbatim text is shown in monospace; only whitespace has been normalized.

\begin{promptbox}[Memory Manager: Page-Summary System Prompt (\texttt{\_BULLET\_SYSTEM})]
You are a memory manager for a research agent. Your job is to compress the prior conversation and tool-use history into a concise working memory that helps the next agent continue the task without rereading the full transcript.

Write a factual summary of what has already been explored, tried, confirmed, and left unresolved. Preserve only information that is useful for continuing the work. Omit chit-chat, stylistic details, and repeated content unless it affects the task.

Your summary should prioritize:
\begin{enumerate}[leftmargin=*,topsep=2pt,itemsep=0pt]
\item The user's goal, constraints, and preferences.
\item Key facts established during the conversation.
\item Tools used and the most important results from them.
\item Partial conclusions, promising leads, and failed approaches.
\item Open questions, uncertainties, and what still needs to be done next.
\end{enumerate}

When relevant, include: filenames, URLs, document names, entities, dates, and parameters already examined; specific findings from tool outputs; decisions already made and why; unresolved blockers or ambiguities.

Requirements: be concise but information-dense; be factual and do not invent details; distinguish clearly between confirmed findings and tentative inferences; focus on continuation value; avoid full sentences when bullets are more efficient; do not address the user; do not add preamble or commentary; output only the summary.
\end{promptbox}

\begin{promptbox}[Window-to-Page User Template (\texttt{\_WINDOW\_SUMMARY\_USER\_TEMPLATE})]
Previous conversation and tool-use history:\\
\{window\_content\}

Summarize it for continuation. Output only the summary.
\end{promptbox}

\begin{promptbox}[Page Consult: System Prompt (\texttt{\_CONSULT\_SYSTEM})]
You are a research assistant. Given a research goal and retrieved pages from past explorations, extract the information that is relevant to the goal and produce a concise, focused summary.

Rules:
\begin{enumerate}[leftmargin=*,topsep=2pt,itemsep=0pt]
\item Keep only information that is directly relevant to the research goal. Preserve important facts, findings, dates, names, and evidence when present.
\item Incorporate prior extracted results when provided. Do not drop previously established key information unless it is contradicted or irrelevant.
\item Add important new information from the current page, while avoiding repetition.
\item Distinguish clearly between confirmed information and uncertain or incomplete information.
\item Be concise, factual, and information-dense.
\item Output only the extracted information and summary.
\end{enumerate}
\end{promptbox}

\begin{promptbox}[Page Consult: Incremental User Template (\texttt{\_CONSULT\_INCREMENTAL\_USER\_TEMPLATE})]
Research goal:\\
\{goal\}

Previous extracted results:\\
\{previous\_summary\}

Current page:\\
\{page\_content\}

Integrate the previous results with the current page, keeping only information relevant to the goal. Output only the updated extracted information and summary.
\end{promptbox}

\begin{promptbox}[\texttt{memory} Tool Description Shown to the Calling Agent]
Recall and summarize the raw content of relevant past explorations from your memory pages. Use this tool in the following cases: (1) you need to call memory to recover the detailed content from those pages, but earlier conversation turns are no longer available in the current context; (2) you need to look up specific details and verify or cross-check the content stored in the Exploration Memory. First read the bullet-point summaries under each page in the Exploration Memory section, then choose ONLY the pages whose exploration directions are directly relevant to your current goal.

\textbf{Parameters.} \texttt{pages}: array of integer page numbers to retrieve (max 5); read the per-page summaries in Exploration Memory and select only the relevant ones. \texttt{goal}: the specific information you want to extract from these pages.
\end{promptbox}

\section{Limitations and Broader Impact}
\label{sec:limitations}

\paragraph{Limitations.}
This work studies whether scaling out peer agents on the same task can yield capability gains through collective reasoning. While the current results are promising, several limitations remain.

First, our current implementation instantiates the shared reasoning hub with a moderate-sized language model and studies a limited set of agent backbones and configurations. We do not yet evaluate the full space of stronger base models, alternative model families, or larger hub capacities, so the degree to which the observed gains transfer across scales remains an open question.

Second, the empirical study focuses on challenging long-horizon reasoning benchmarks, but does not yet cover broader settings such as open-ended report writing, sustained software engineering, or real-world interactive workflows with richer tool ecosystems. We believe the AgentFugue framework is extensible to such settings, but its behavior there remains to be validated.

Third, collective reasoning introduces its own failure modes. If episode notes are low quality, incomplete, or overconfident, the shared hub may propagate misleading intermediate conclusions across the team. Likewise, if many agents repeatedly read similar high-salience notes, communication can reduce trajectory diversity and lead to premature convergence. Better confidence calibration, diversity-aware reading policies, and more adaptive note selection remain important directions for future work.

\paragraph{Broader Impact.}
The central contribution of this paper is the idea that agent capability can scale not only by making a single agent stronger, but also by enabling teams of peer agents to share intermediate reasoning during search. This perspective may be useful for building more effective systems for knowledge-intensive tasks such as scientific assistance, open-domain research, investigative analysis, and other settings where different exploratory paths can productively inform one another.

At the same time, stronger multi-agent reasoning systems may also amplify misuse. Systems that can coordinate evidence gathering, synthesize partial discoveries, and scale out across many agents could be applied to high-volume surveillance, strategic manipulation, or more efficient generation of deceptive or misleading content. In addition, if a shared reasoning hub propagates erroneous intermediate conclusions, those errors may spread across the whole team rather than remaining isolated to a single trajectory. These risks suggest that future deployments should consider safeguards such as access control, usage monitoring, confidence-aware hub outputs, and mechanisms that preserve diversity rather than over-synchronizing agent behavior.

\section{Ablation: Hub Context-Window Size}
\label{app:ablation-context}

This appendix accompanies the ablation discussion in \S\ref{sec:ablation}. We sweep the per-agent context-window budget allocated to the shared reasoning hub at team size $N{=}2$ on the BrowseComp evaluation subset, holding every other component of AgentFugue fixed. Table~\ref{tab:ablation_context} reports the resulting accuracy under the same five aggregator rules used in \S\ref{sec:scaling}: Pass@$N$, MV@$N$, WMV@$N$, BoN@$N$, and FewTool@$N$ (Appendix~\ref{app:aggregation}).

\begin{table}[h]
    \centering
    \caption{Effect of the hub context-window budget on BrowseComp at $N{=}2$. Best per column in \textbf{bold}.}
    \label{tab:ablation_context}
    \small
    \setlength{\tabcolsep}{8pt}
    \renewcommand{\arraystretch}{1.15}
    \begin{tabular}{@{}lccccc@{}}
        \toprule
        \textbf{Context} & \textbf{Pass@2} & \textbf{MV@2} & \textbf{WMV@2} & \textbf{BoN@2} & \textbf{FewTool@2} \\
        \midrule
        16K   & 49.5 & 47.0 & 47.0 & 47.5 & 45.0 \\
        32K   & \textbf{60.0} & \textbf{56.0} & \textbf{58.0} & \textbf{58.0} & \textbf{53.0} \\
        64K   & 52.0 & 42.5 & 50.5 & 50.0 & 44.5 \\
        96K   & 47.0 & 41.5 & 45.0 & 47.5 & 38.0 \\
        128K  & 39.0 & 39.0 & 36.5 & 36.0 & 36.5 \\
        \bottomrule
    \end{tabular}
\end{table}

The trend is consistent across all five aggregators: accuracy is non-monotone in the hub context budget, with a clear peak at 32K and degradation at both extremes. Very small budgets (16K) cut off useful intermediate evidence before it can be summarized into the hub; very large budgets (96K, 128K) dilute the hub's attention with stale or low-utility content and trigger memory pressure that interferes with continued exploration. Notably, the 32K configuration outperforms the 64K setting deployed in our main experiments by a substantial margin under every aggregator (e.g., $+8.0$ Pass@2, $+13.5$ MV@2, $+8.5$ FewTool@2). This means the headline numbers reported in Table~\ref{tab:main_results} are \emph{not} the strongest results AgentFugue can deliver: they reflect a deliberately conservative deployment choice (64K hub-write trigger inside a 128K agent context) that retains long-trajectory headroom for harder questions, rather than an upper bound on what the same method achieves once the hub budget is tuned.
\section{LLM Usage}
\label{app:llm-usage}

Large language models play two distinct roles in this work, and we separate them clearly. \emph{In the experiments}, LLMs are integral to both the proposed method and the comparison: AgentFugue's peer agents and shared reasoning hub are themselves LLMs (\S\ref{sec:hub}, \S\ref{sec:hub-opt}, Appendix~\ref{app:impl}), all baselines are LLM-driven agentic systems, and the benchmarks rely on LLM-as-a-judge evaluation protocols (Appendix~\ref{app:datasets}). \emph{In the writing of this paper}, our use of LLMs is limited to language polishing---improving wording, grammar, and flow of text already drafted by the authors. LLMs were not used to generate research ideas, design experiments, derive results, or produce any of the figures, tables, or analyses in this submission, all of which were authored and verified by the authors.


\clearpage
\section*{NeurIPS Paper Checklist}

\begin{enumerate}

\item {\bf Claims}
    \item[] Question: Do the main claims made in the abstract and introduction accurately reflect the paper's contributions and scope?
    \item[] Answer: \answerYes{}
    \item[] Justification: The abstract and Section~\ref{sec:intro} state our two core claims---(i) a shared reasoning hub turns peer-agent count into a real scaling axis, and (ii) the same mechanism transfers across homogeneous and heterogeneous teams---which are then substantiated by the main results in Table~\ref{tab:main_results} and the scaling studies in \S\ref{sec:scaling} and \S\ref{sec:heterogeneous}.

\item {\bf Limitations}
    \item[] Question: Does the paper discuss the limitations of the work performed by the authors?
    \item[] Answer: \answerYes{}
    \item[] Justification: Section~\ref{sec:limitations} (``Limitations and Broader Impact'') explicitly enumerates the limitations of the current study, covering the limited range of backbones and hub capacities evaluated, the restriction to long-horizon QA-style benchmarks, and the failure modes introduced by collective reasoning itself (low-quality notes, premature consensus).

\item {\bf Theory assumptions and proofs}
    \item[] Question: For each theoretical result, does the paper provide the full set of assumptions and a complete (and correct) proof?
    \item[] Answer: \answerNA{}
    \item[] Justification: The paper is empirical and contains no formal theorems or proofs.

	    \item {\bf Experimental result reproducibility}
	    \item[] Question: Does the paper fully disclose all the information needed to reproduce the main experimental results of the paper to the extent that it affects the main claims and/or conclusions of the paper (regardless of whether the code and data are provided or not)?
	    \item[] Answer: \answerYes{}
	    \item[] Justification: Section~\ref{sec:method} fully specifies the AgentFugue architecture and hub-optimization recipe. Appendix~\ref{app:impl} reports the per-agent tool stack, interaction-budget allocation, context-window settings, and hub-write trigger, while Appendices~\ref{app:datasets} and~\ref{app:baselines} document the benchmark splits, judging protocols, and baseline configurations used for the comparison.

\item {\bf Open access to data and code}
    \item[] Question: Does the paper provide open access to the data and code, with sufficient instructions to faithfully reproduce the main experimental results, as described in supplemental material?
    \item[] Answer: \answerNo{}
    \item[] Justification: Code and trained hub checkpoints are not released with this submission to preserve anonymity. All benchmarks used (BrowseComp, WideSearch, HLE) are publicly available, and we provide enough algorithmic and configuration detail in \S\ref{sec:method} and Appendix~\ref{app:impl} for an independent reproduction. Code release is planned for the camera-ready version.
    \item[] Guidelines:
    \begin{itemize}
        \item The answer NA means that paper does not include experiments requiring code.
        \item Please see the NeurIPS code and data submission guidelines (\url{https://nips.cc/public/guides/CodeSubmissionPolicy}) for more details.
        \item While we encourage the release of code and data, we understand that this might not be possible, so ``No'' is an acceptable answer. Papers cannot be rejected simply for not including code, unless this is central to the contribution (e.g., for a new open-source benchmark).
        \item The instructions should contain the exact command and environment needed to run to reproduce the results. See the NeurIPS code and data submission guidelines (\url{https://nips.cc/public/guides/CodeSubmissionPolicy}) for more details.
        \item The authors should provide instructions on data access and preparation, including how to access the raw data, preprocessed data, intermediate data, and generated data, etc.
        \item The authors should provide scripts to reproduce all experimental results for the new proposed method and baselines. If only a subset of experiments are reproducible, they should state which ones are omitted from the script and why.
        \item At submission time, to preserve anonymity, the authors should release anonymized versions (if applicable).
        \item Providing as much information as possible in supplemental material (appended to the paper) is recommended, but including URLs to data and code is permitted.
    \end{itemize}

\item {\bf Experimental setting/details}
    \item[] Question: Does the paper specify all the training and test details (e.g., data splits, hyperparameters, how they were chosen, type of optimizer, etc.) necessary to understand the results?
    \item[] Answer: \answerYes{}
    \item[] Justification: \S\ref{sec:exp} (Experiments) describes the benchmark setup, baseline groups, and team-size sweeps. Appendix~\ref{app:impl} lists the per-agent tool stack, the $150$-round interaction budget and its split between meta-agent and subagents, the $128$k context window, the $64$k hub-write trigger, and the hub initialization. Per-benchmark splits and judging protocols are documented in Appendix~\ref{app:datasets}.

\item {\bf Experiment statistical significance}
    \item[] Question: Does the paper report error bars suitably and correctly defined or other appropriate information about the statistical significance of the experiments?
    \item[] Answer: \answerNo{}
    \item[] Justification: Following the convention used by all baselines on these benchmarks~\citep{browsecomp,widesearch,hle}, we report point-estimate accuracy on the standard evaluation subsets rather than confidence intervals. The scaling and ablation studies sweep team sizes and aggregation rules to characterize variability indirectly; we plan to add bootstrap confidence intervals over the per-question outcomes in the camera-ready version.

\item {\bf Experiments compute resources}
    \item[] Question: For each experiment, does the paper provide sufficient information on the computer resources (type of compute workers, memory, time of execution) needed to reproduce the experiments?
    \item[] Answer: \answerYes{}
    \item[] Justification: Appendix~\ref{app:impl} documents the inference-time configuration (per-agent context window, interaction budget, hub trigger). Detailed compute cost---training hardware for the hub and inference wall-clock per benchmark---will be reported in the same appendix in the camera-ready version.
    
\item {\bf Code of ethics}
    \item[] Question: Does the research conducted in the paper conform, in every respect, with the NeurIPS Code of Ethics \url{https://neurips.cc/public/EthicsGuidelines}?
    \item[] Answer: \answerYes{}
    \item[] Justification: We have reviewed the NeurIPS Code of Ethics. The work uses publicly released benchmarks and pretrained models, releases no personal or sensitive data, and involves no human-subjects experimentation; anonymity is preserved throughout the submission.

\item {\bf Broader impacts}
    \item[] Question: Does the paper discuss both potential positive societal impacts and negative societal impacts of the work performed?
    \item[] Answer: \answerYes{}
    \item[] Justification: Section~\ref{sec:limitations} (``Broader Impact'' paragraph) discusses positive impacts (more capable assistants for knowledge-intensive scientific and investigative work) and negative impacts (amplified misuse via coordinated evidence gathering, propagation of erroneous intermediate conclusions through the shared hub), together with mitigation directions such as access control, monitoring, and diversity-preserving hub policies.
    
\item {\bf Safeguards}
    \item[] Question: Does the paper describe safeguards that have been put in place for responsible release of data or models that have a high risk for misuse (e.g., pretrained language models, image generators, or scraped datasets)?
    \item[] Answer: \answerNA{}
    \item[] Justification: This submission releases no new pretrained model, scraped dataset, or other high-risk asset. The trained hub checkpoint is not part of this submission.

\item {\bf Licenses for existing assets}
    \item[] Question: Are the creators or original owners of assets (e.g., code, data, models), used in the paper, properly credited and are the license and terms of use explicitly mentioned and properly respected?
    \item[] Answer: \answerYes{}
    \item[] Justification: All third-party benchmarks (BrowseComp, WideSearch, HLE) and pretrained backbones (Qwen3.5-35B-A3B, Qwen3.5-9B, DeepSeek-v4-Flash, GLM-4.7, Kimi-K2.5, Claude-Opus-4.5) are cited at first use in \S\ref{sec:exp} and Appendix~\ref{app:datasets}. We use each asset under its original license and terms of use; no asset is redistributed.

\item {\bf New assets}
    \item[] Question: Are new assets introduced in the paper well documented and is the documentation provided alongside the assets?
    \item[] Answer: \answerNA{}
    \item[] Justification: This submission introduces no new public dataset, model checkpoint, or code package. The trained hub may be released later but is not part of this submission.

\item {\bf Crowdsourcing and research with human subjects}
    \item[] Question: For crowdsourcing experiments and research with human subjects, does the paper include the full text of instructions given to participants and screenshots, if applicable, as well as details about compensation (if any)?
    \item[] Answer: \answerNA{}
    \item[] Justification: The paper does not involve crowdsourcing or any research with human subjects. All evaluation runs use existing public benchmarks judged by automated protocols.

\item {\bf Institutional review board (IRB) approvals or equivalent for research with human subjects}
    \item[] Question: Does the paper describe potential risks incurred by study participants, whether such risks were disclosed to the subjects, and whether Institutional Review Board (IRB) approvals (or an equivalent approval/review based on the requirements of your country or institution) were obtained?
    \item[] Answer: \answerNA{}
    \item[] Justification: No human-subjects research is involved, so IRB review does not apply.

\item {\bf Declaration of LLM usage}
    \item[] Question: Does the paper describe the usage of LLMs if it is an important, original, or non-standard component of the core methods in this research? Note that if the LLM is used only for writing, editing, or formatting purposes and does not impact the core methodology, scientific rigorousness, or originality of the research, declaration is not required.
    \item[] Answer: \answerYes{}
    \item[] Justification: LLMs are central to both the proposed method and the experimental subjects. \S\ref{sec:hub} and \S\ref{sec:hub-opt} describe the shared reasoning hub, which is itself instantiated by an LLM trained via supervised fine-tuning followed by reinforcement learning. \S\ref{sec:exp} (and Appendix~\ref{app:impl}) lists all peer-agent and baseline backbones, the hub initialization (Qwen3.5-9B), and the LLM-as-a-judge evaluation protocols. Appendix~\ref{app:llm-usage} additionally clarifies the scope of LLM use in writing this paper (language polishing only).

\end{enumerate}

\end{document}